# Optimizing Automatic Summarization of Long Clinical Records Using Dynamic Context Extension Testing and Evaluation of the NBCE Method

Guoqing ZHANG[†], Keita FUKUYAMA[††], Kazumasa KISHIMOTO[††], and Tomohiro KURODA [††]

[†]Social Informatics, Kyoto University
Sakyo-ku, Kyoto 606-8501 JAPAN

[††]Medical Informatics, Kyoto University Hospital
Sakyo-ku, Kyoto 606-8501 JAPAN

54 Kawahara-cho, Shogoin, Sakyo-ku, Kyoto 606-8507, Japan
E-mail: [†]kuochingcha@gmail.com, [††]{fkeita, kishimoto, tomo}@kuhp.kyoto-u.ac.jp

**Abstract** Summarizing patient clinical notes is vital for reducing documentation burdens. Current manual summarization makes medical staff struggle. We propose an automatic method using LLMs, but long inputs cause LLMs to lose context, reducing output quality especially in small size model. We used a 7B model, open-calm-7b, enhanced with Native Bayes Context Extend and a redesigned decoding mechanism to reference one sentence at a time, keeping inputs within context windows, 2048 tokens. Our improved model achieved near-parity with Google's over 175B Gemini on ROUGE-L metrics with 200 samples, indicating strong performance using less resources, enhancing automated EMR summarization feasibility.

## 1. Introduction

In clinical practice, summarizing patient clinical records, commonly known as patient summaries, is valuable for quickly understanding patient information. However, since a significant portion of healthcare professionals' working hours is devoted to creating these summaries, there is a need to improve the efficiency and automation of this process. With the recent advancements in large language models (LLMs), it has become increasingly possible to automate complex tasks involving natural language. Text summarization using LLMs has shown practical performance in summarizing academic papers and news content, raising expectations for its application in generating summaries within healthcare settings.

However, when implementing automated summarization of clinical records using LLMs, the limitations of the context window become a challenge. While individual clinical entries may not be particularly lengthy, the accumulated records over extended patient treatment periods can eventually exceed the maximum token limit that each LLM can handle, significantly surpassing the context window capacity. According to research by Rewon Child, the memory and computational requirements of the core formula of Transformers, Attention, increase proportionally to the square of the sequence length [1]. As the context window length increases, the amount of input information the model must process also grows, leading to an increase in the number of model parameters and computational complexity, thereby enlarging the model size. Specifically, in

LLMs, the Attention mechanism needs to compute the relationships between every word and all other words within the input sequence. The computational load for this process increases exponentially with the length of the context window. Consequently, as the number of model parameters grows, so does the need for storage and computational resources, such as GPU memory. When inputs exceed the context window set during training, the combination of long-term Attention deficiency in the Attention algorithm leads to catastrophic forgetting when LLMs process long contexts, resulting in a decline in output quality [2][3][4].

In the case of recent cloud-based LLMs with over 200B parameters, the maximum token capacity has also grown significantly, enabling the summarization of long texts, like Gemini1.5 pro can accept 1,000,000 tokens. However, when creating summaries of long clinical texts in healthcare settings, the nature of the data imposes certain restrictions on the use of cloud-based LLMs. On the other hand, the use of LLMs on-premises is limited by GPU memory size, and it is not easy to introduce expensive hardware capable of handling models exceeding 100B parameters. By solving this issue with relatively smaller on-premises LLMs, ranging from 7B to 30B in scale, driven by smaller GPUs within electronic medical record (EMR) networks, it is expected that a summarization system that can be promptly deployed in real-world clinical settings can be realized.

**2. Related Works**

ClinicalBERT is known for its research on using language models to analyze clinical data [5]. This research focuses on analyzing time-series data within hospital information systems to predict the likelihood of patient readmission after discharge at a specific point in time, but it does not produce Text-to-Text outputs. Following the introduction of ChatGPT, research on the automatic summarization of clinical records using large language models has been reported by Dave Van Veen [6]. Using the Open-i radiology report dataset, he evaluated current mainstream open-source and closed-source models using methods like BLEU and ROUGE-L, conducting multiple evaluations with different prompts and hyperparameters. The best result reported was a ROUGE-L (F1) score of 35.5%.

In terms of summarizing Japanese medical texts, there are reports utilizing case report articles, but these mostly use relatively short and well-structured texts [7]. In contrast, clinical records in healthcare settings include significantly longer texts compared to radiology reports or case studies. Nakagawa et al. reported a system that extracts necessary elements by selecting them after Named Entity Recognition (NER) and then generates text. However, due to the selection process post-NER, the summary generation itself is not fully automated [8].

Regarding research on extending the context window, two main approaches are highlighted: Parallel Context Window (PCW) [9] and Naive Bayes-based Context Extension (NBCE) [10]. Su Jianlin's paper compares NBCE and PCW, showing that NBCE achieves superior results across multiple benchmarks.

**3. Method**

We enumerated all combinations and added prompt information to create a single training instance, fine-tuning the Open-Calm-7B model, maxing input length is 2048 tokens.

To address the issue of losing context due to long inputs, we applied Dr. Su Jianlin's NBCE method, which significantly extends the context window limit by splitting long context and processing them individually as inputs to the LLM. Tokens from each output were selected to form the final response. This approach involved modifying the LLM's decoding layer to decode reference texts in parallel along with prompts. Using information entropy as an indicator, the decoder selects output tokens dynamically.

In terms of implementation, when the model receives all clinical records for a single patient, an algorithm samples the data set with a sampling rate of 0.15 based on the distribution of token lengths, and then performs parallel decoding on the sampled dataset.

During decoding, Shannon's information entropy was used. When generating the next token, our algorithm independently computes the Shannon information entropy of all preceding inputs and selects the preceding input with the lowest entropy as the basis for the conditional probability distribution of the next token to fetch the most adoptive reference content for next token predicting, miniest entropy strategy. This process is repeated for each generated token until the predetermined maximum number of tokens is reached (Figure 2: Structure of NBCE decoding processing. The medical record entries required for summary generation (such as S, O, current medical history, etc. on the left side) will be input for each individual medical record. The token will be estimated for each input).

$$Shannon\ Entropy = -\sum_{i=1}^{n} p(token_i) \log(p(token_i))$$

*n is tokens number from one EHR sentence, $p(token_i)$ is the posibility of $token_i$*

Through our investigation of actual clinical records, we observed that the length of clinical notes exhibits a high variance. This variability may obscure the patterns between clinical notes and their corresponding summaries, making it challenging for large language models to learn effectively. However, manually denoising the dataset is time-consuming and costly. Therefore, we propose a method to denoise the training set as efficiently as possible.

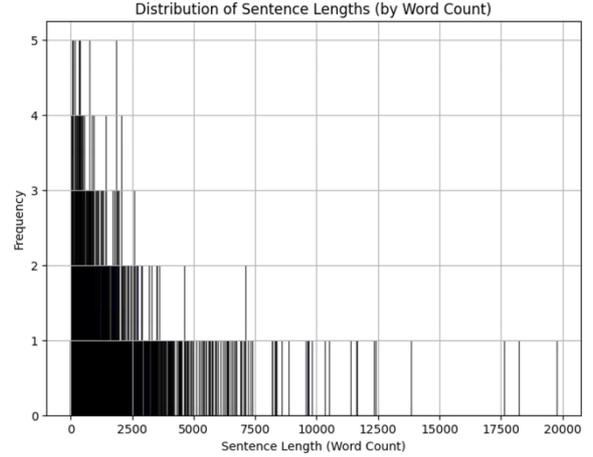

Figure 1: Clinical Notes length distribution

| Samples | 228 |
|---|---|
| Variance | 2,837,889 |
| Mean | 1,427 |
| Standard Deviation | 1,684 |

Table 1: Variance, Mean, Standard Deviation

We utilized the TF-IDF (Term Frequency-Inverse Document Frequency) method to convert each clinical note into a vector representation, applying the same procedure to the summaries. Subsequently, we computed the cosine similarity between each clinical note vector and all summary vectors, with values ranging from 0 to 1. A result closer to 1 indicates a higher similarity between two records, while values closer to 0 indicate dissimilarity.

$$tf_{i,j} = \frac{n_{i,j}}{\sum_k n_{k,j}}$$

$$idf_i = \log \frac{|D|}{|\{j: t_i \in d_j\}|}$$

In the above formula, suppose $d_j$ sentence contains $k$ words, where $n_{k,j}$ represents the number of times the term $t_k$ appears in sentence $d_j$. Therefore, the numerator $n_{i,j}$ is the frequency of the term $t_i$ in sentence $d_j$, while the denominator is the sum of the frequencies of all terms in sentence $d_j$. $|D|$ is number of documents, $|\{j: t_i \in d_j\}|$ is the number of sentences containing the term $t_i$. If the term does not appear in any documents, it results in a denominator of zero.

Then,
$$tfidf_{i,j} = tf_{i,j} \times idf_i$$

For cosine similarity:

$Similarity$
$$= \frac{Record \cdot Summary}{\| Record \| \| Summary \|}$$
$$= \frac{\sum_1^n tfidf_{Record} tfidf_{Summary}}{\sqrt{\sum_1^n (tfidf_{Record})^2} \times \sqrt{\sum_1^n (tfidf_{Summary})^2}}$$

Finally, we fine-tuned the Open-Calm-7B model using the LoRA (Low-Rank Adaptation) method.

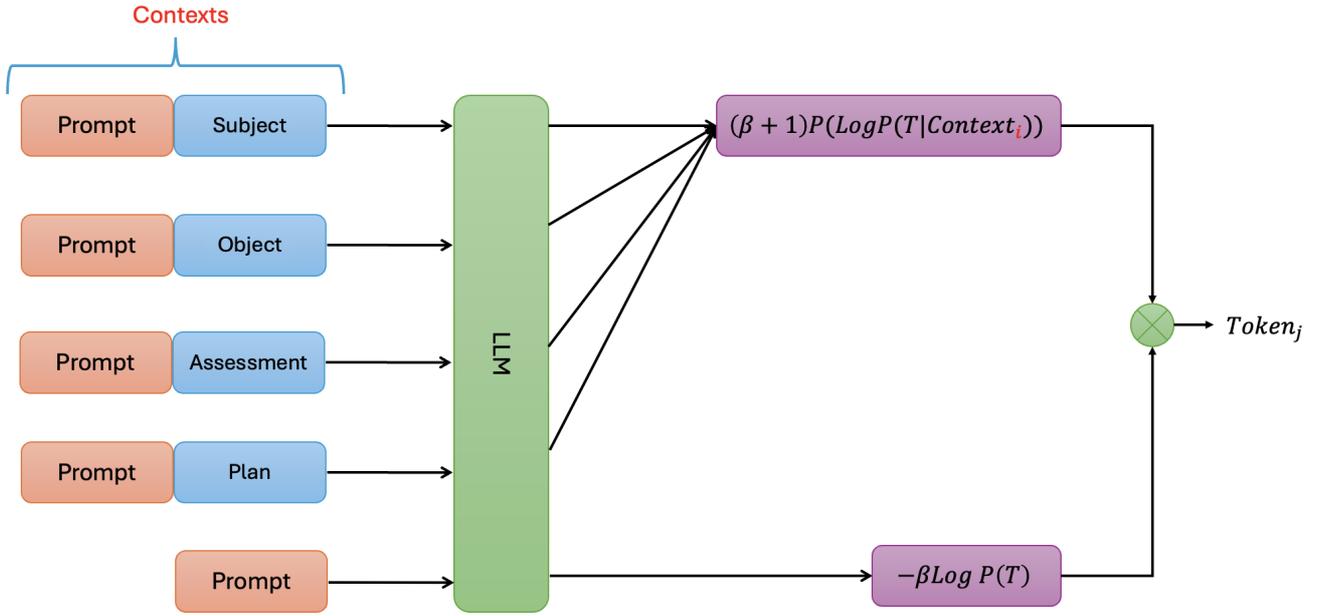

Figure 2: Structure of NBCE decoding processing. The medical record entries required for summary generation (such as S, O, current medical history, etc. on the left side) will be input for each individual medical record. The token will be estimated for each input

## 4. Experiment

### 4.1 Dataset and Preprocessing

For data collection, we sourced patient records from the electronic medical records database of Kyoto University Hospital, covering nearly a decade from 2013 to 2023. This dataset includes approximately 500,000 samples from around 5,000 patients. We calculated the cosine similarity for all samples, sorted them in descending order, and selected the top 160,000 records as the training set. From this set, we randomly sampled 2,000 records as the validation dataset and 200 records as the test dataset.

### 4.2 Evaluation Metrics

The generated summary was evaluated using the ROUGE-L (Recall-Oriented Understudy for Gisting Evaluation-Longest Common Subsequence) score, with the summary created by physicians as the ground truth data. ROUGE-L is one of the metrics used to assess the similarity between the generated text and the reference text. It is primarily applied in the evaluation of automatic summarization, measuring text similarity based on the Longest Common Subsequence (LCS). The calculation of ROUGE-L is represented by the following formula:

$$ROUGE - L = \frac{LCS(generated, Human\ Write)}{Length\ of\ Human\ Write}$$

Longest Common Subsequence (LCS): ROUGE-

L is based on the LCS between the generated text and the reference text. The LCS refers to the longest subsequence that appears in both sequences without rearranging the order of characters.

Precision, Recall, F1 Score: ROUGE-L calculates precision (using the model output as the reference), recall (using the physician-written summary as the reference), and the F1 score (the harmonic mean of precision and recall).

Applications: ROUGE-L is used to evaluate the quality of text in tasks such as text summarization, machine translation, and dialogue generation. It is particularly useful for assessing the similarity of longer texts, as it can capture relatively long matching sequences.

### 4.3 Baseline

We use Google's Gemini as baseline model which size is over 175B and maximum acceptable input is 1,000,000 tokens, do this Comparison experiment. Open-Calm-7B we use is one of its 25ths, and the maximum acceptable input is 2048, after tuning on 160,000 samples.

### 4.4. Result

Here are the ROUGE-L scores evaluated for 200 summaries across Human Writing, our model (Figure 3), and Google Gemini (Figure 4), average score in Table 2.

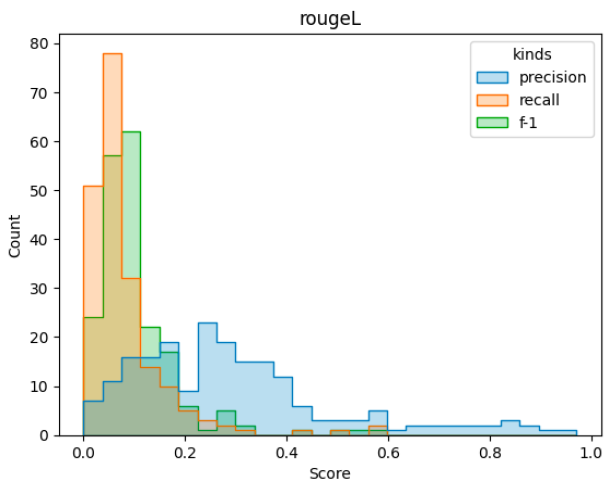

Figure 3: Distribution of ROUGE-L Precision, Recall, and F1 scores across individual samples by Open-Calm-7B with NBCE

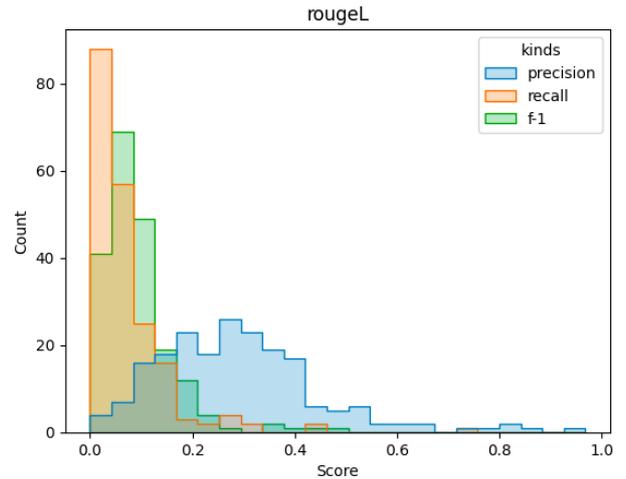

Figure 4: Distribution of ROUGE-L Precision, Recall, and F1 scores across individual samples by Gemini

| ROUGE-L | Open-Calm-7B sample rate (0.15) | Gemini sample rate (1.0) |
|---|---|---|
| Precision | **0.2954** | 0.2277 |
| Recall | 0.0824 | 0.1508 |
| F-1 | 0.1043 | 0.1473 |

Table 2: Comparison of average ROUGE-L scores between Open-Calm-7B (using NBCE with a 0.15 sampling rate) and Gemini (full sampling) on a test set of 200 samples.

### 5. Discussion

In this study, we proposed an optimization strategy that enables a small language model with 7 billion parameters to achieve performance comparable to a commercial model with 175 billion parameters on medical clinical summarization tasks. This breakthrough offers significant practical value, especially in healthcare settings, by substantially reducing costs while enhancing data security.

1. Cost Efficiency and Scalability of Small Models. Compared to large language models, small models (7B) offer significant advantages in terms of deployment costs. Large models typically require expensive computational resources, such as high-performance GPU clusters, as well as substantial memory and storage capacity, which can limit their deployment in practical applications. Our approach enables the 7B model to deliver high-quality medical summarization output even in environments with limited hardware resources. This cost-effective nature makes small models more suitable for large-scale deployment in hospitals, not only reducing hardware expenses but also lowering ongoing operational costs associated with cloud computing. As hospitals increasingly demand intelligent healthcare solutions, the affordability of these models can facilitate wider adoption, thereby enhancing the overall intelligence of healthcare systems.

2. Enhancing Data Security and Privacy through Local Deployment. Data security and patient privacy are critical considerations in medical applications. Traditional cloud-based AI models often require patient data to be uploaded to remote servers, which poses risks of data breaches and cyberattacks. Our proposed small model can be fully deployed within a hospital's local network, effectively eliminating the risk of data leakage during transmission. By processing data entirely on-premises, sensitive patient information is better protected, ensuring compliance with stringent data privacy regulations (e.g., GDPR, HIPAA). This localized data processing approach offers healthcare institutions a secure way to leverage AI technologies while safeguarding patient confidentiality.

3. Reducing Communication Latency to Improve Clinical Decision-Making. In many clinical scenarios that require rapid response, such as emergency rooms or operating theaters, the speed of model inference directly impacts the timeliness of medical decisions. Cloud-based large models often suffer from network latency due to the need for remote communication, which can lead to delays in critical situations. In contrast, our small model, deployed locally, virtually eliminates communication latency, significantly enhancing response times. This low-latency characteristic makes local models particularly suitable for real-time decision-making in healthcare settings, further improving their clinical utility.

## 6. References

[1]. Child, R., Gray, S., Radford, A., & Sutskever, I. (2019). Generating long sequences with sparse transformers. arXiv preprint arXiv:1904.10509.

[2]. Beltagy, I., Peters, M. E., & Cohan, A. (2020). Longformer: The long-document transformer. arXiv preprint arXiv:2004.05150.

[3]. Brown, T. B. (2020). Language models are few-shot learners. arXiv preprint arXiv:2005.14165.

[4]. Wolf, T., et al. "TransferTransfo: A Transfer Learning Approach for Neural Chatbots." arXiv preprint arXiv:1901.08149, 2019.

[5]. Huang, K., Altosaar, J., & Ranganath, R. (2019). Clinicalbert: Modeling clinical notes and predicting hospital readmission. arXiv preprint arXiv:1904.05342.

[6]. Van Veen, D., Van Uden, C., Blankemeier, L., Delbrouck, J. B., Aali, A., Bluethgen, C., ... & Chaudhari, A. S. (2024). Adapted large language models can outperform medical experts in clinical text summarization. Nature medicine, 30(4), 1134-1142.

[7]. 清水 聖司, 矢田 竣太郎, 荒牧 英治：所望の患者データを作る：Variational Auto-Encoder による症例報告生成, 言語処理学会 第29回年次大会（NLP2023）発表論文集, H11-2, pp. 2731-2736, 2023（2023/3/16, 沖縄コンベンションセンター

[8]. https://www.jstage.jst.go.jp/article/pjsai/JSAI2024/0/JSAI2024_3S5OS7c02/_article/-char/ja

[9]. Ratner, N., Levine, Y., Belinkov, Y., Ram, O., Magar, I., Abend, O., ... & Shoham, Y. (2022). Parallel context windows for large language models. arXiv preprint arXiv:2212.10947.

[10]. Su, J., Ahmed, M., Ao, L., Zhu, M., & Liu, Y. (2024). Naive Bayes-based Context Extension for Large Language Models. arXiv preprint arXiv:2403.17552.